\PassOptionsToPackage{unicode}{hyperref}
\PassOptionsToPackage{hyphens}{url}
\PassOptionsToPackage{dvipsnames,svgnames,x11names}{xcolor}
\documentclass[
  11pt,
]{article}
\usepackage{xcolor}
\usepackage[margin=1in]{geometry}
\usepackage{amsmath,amssymb}
\usepackage{iftex}
\ifPDFTeX
  \usepackage[T1]{fontenc}
  \usepackage[utf8]{inputenc}
  \usepackage{textcomp} 
\else 
  \usepackage{unicode-math} 
  \defaultfontfeatures{Scale=MatchLowercase}
  \defaultfontfeatures[\rmfamily]{Ligatures=TeX,Scale=1}
\fi
\usepackage{lmodern}
\ifPDFTeX\else
\fi
\IfFileExists{upquote.sty}{\usepackage{upquote}}{}
\IfFileExists{microtype.sty}{
  \usepackage[]{microtype}
  \UseMicrotypeSet[protrusion]{basicmath} 
}{}
\makeatletter
\@ifundefined{KOMAClassName}{
  \IfFileExists{parskip.sty}{%
    \usepackage{parskip}
  }{
    \setlength{\parindent}{0pt}
    \setlength{\parskip}{6pt plus 2pt minus 1pt}}
}{
  \KOMAoptions{parskip=half}}
\makeatother
\usepackage{longtable,booktabs,array}
\usepackage{calc} 
\usepackage{etoolbox}
\makeatletter
\patchcmd\longtable{\par}{\if@noskipsec\mbox{}\fi\par}{}{}
\makeatother
\IfFileExists{footnotehyper.sty}{\usepackage{footnotehyper}}{\usepackage{footnote}}
\makesavenoteenv{longtable}
\providecommand{\tightlist}{%
  \setlength{\itemsep}{0pt}\setlength{\parskip}{0pt}}
\usepackage{bookmark}
\IfFileExists{xurl.sty}{\usepackage{xurl}}{} 
\hypersetup{
  pdftitle={Training a Language Model End-to-End in Rust: An Experience Report},
  pdfauthor={Arif Ahmed Adito (Adioris Tech Ltd., Dhaka, Bangladesh)},
  colorlinks=true,
  linkcolor={Maroon},
  filecolor={Maroon},
  citecolor={Blue},
  urlcolor={Blue},
  pdfcreator={LaTeX via pandoc}}

\title{Training a Language Model End-to-End in Rust: An Experience Report}
\author{Arif Ahmed Adito (Adioris Tech Ltd., Dhaka, Bangladesh)}
\date{}

\begin{document}
\maketitle

\subsection{Abstract}\label{abstract}

I pretrained a language model end-to-end in Rust --- alone, with no team, no PyTorch, and no Python anywhere in the training path --- for \textbf{\$164} in rented GPU time. I report that as an achievement, not a recommendation: the more useful contribution of this paper is a \textbf{measured failure taxonomy} of the two leading Rust ML frameworks, Candle and Burn, as \emph{training} (not inference) backends in 2026 --- five distinct Candle defects, including fused kernels that silently produce no gradient at all, and three Burn defects, including a backward pass I measured at roughly 3\% of theoretical GPU throughput and a kernel-fusion path that segfaults mid-training at multi-billion-parameter scale. Every one of these defects passed ordinary loss-curve inspection; none of them announced itself. I describe the verification discipline that caught six such silent failures before they could waste the compute budget, centered on a \textbf{gradient-flow arbiter}: a test that runs one forward/backward pass and asserts every trainable parameter receives a finite, nonzero gradient, generalizable to any framework. The trained model (roughly 0.4B parameters, Bangla-first) shows strong Bangla language-modeling signal --- a per-token negative log-likelihood of 0.93 against 12.60 for a random-initialized twin --- while scoring at chance on English commonsense multiple-choice, the declared and expected outcome of a deliberately small, Bangla-weighted training budget (about 2 billion tokens, 54.6 hours, one rented H100). I also report a tokenizer-fertility trap specific to Bengali script: naive byte-level tokenization collapsed Bangla to roughly 1.4 characters per token against English's 3.9, silently inverting a ``Bangla-first'' corpus's actual language balance; fixing it reached roughly 4.1 characters per token. To my knowledge, this is among the first documented end-to-end LM pretraining runs in pure Rust, though I make no stronger claim than that, and I did not exhaustively search for prior ones. I close on the project's actual trajectory: after this run, I moved model training to PyTorch and kept Rust for on-device serving. I present that pivot as the paper's central finding, not a failure to disclose --- as of this writing, in my hands, Rust is not yet a competitive place to \emph{train} a language model, though it may be a good place to \emph{serve} one.

\begin{center}\rule{0.5\linewidth}{0.5pt}\end{center}

\subsection{1. Motivation}\label{motivation}

Bangladesh has roughly 170 million Bangla speakers and, at the time of this project, no widely-deployed, on-device, Bangla-first language model built by a local team. Renting a general-purpose model over a network connection --- one that is not universally reliable outside Dhaka --- in a language it was not optimized for is the default option. I wanted to test a different one: build small, cheap, and sovereign, and make the whole stack --- tokenizer, training loop, weights, deployment --- something a single engineer could own end-to-end, without depending on a foreign vendor's API or a foreign country's cloud-hosted checkpoint for a model to work at all.

Two considerations pushed me toward Rust specifically, rather than the default PyTorch/CUDA recipe:

\begin{itemize}
\tightlist
\item
  \textbf{Deployment reality.} The serving target is offline, on-device inference on hardware Bangladeshi users already own --- laptops and, eventually, phones --- with no GPU and no persistent network connection assumed. A single memory-safe binary with no Python runtime and no dependency tree to reproduce is a real advantage there, and Rust inference is mature enough to deliver it.
\item
  \textbf{A bet on symmetry.} If Rust could serve the model, could it also train it? A single-language, single-toolchain stack --- tokenizer, sharding, training loop, checkpointing, and inference all in one codebase, compiled once --- is a genuinely appealing systems property. This paper is the record of testing that bet honestly, including where it did not hold, and doing so as a single person rather than a team: the solo-plus-\$164 combination is itself part of what I am reporting, not incidental to it.
\end{itemize}

I did not set out to make a state-of-the-art claim, and I do not make one here. I set out to answer a narrower, more useful question for anyone else considering the same bet: \emph{as of 2026, what actually breaks when you try to pretrain a transformer-scale language model in Rust, and what does it cost --- in dollars and in engineering hours --- to get a real run out the other side?} The rest of this paper is that answer.

\begin{center}\rule{0.5\linewidth}{0.5pt}\end{center}

\subsection{2. The Stack, at a High Level}\label{the-stack-at-a-high-level}

The system I built is a pure-Rust pipeline: a training loop, a data-loading and sharding pipeline, checkpointing with a tested exact-resume guarantee (weights and optimizer state both round-trip through a save/reload cycle without the resumed run diverging from an equivalent continuous one), and CPU/GPU kernel dispatch that runs the same code path on a Mac laptop, a Windows desktop, and a rented Linux GPU box. I built two independent implementations against two different Rust ML frameworks (Candle and Burn) to be able to compare them directly; the pretraining run reported in this paper used the Candle implementation, against a small set of patches I wrote and pinned to a specific fork (§9). I am deliberately not disclosing the internal architecture of this pipeline --- crate layout, file structure, or kernel-dispatch implementation --- in this paper; what follows is the part I think is useful to anyone else, independent of my specific recipe: what actually broke, and how I caught it.

\begin{center}\rule{0.5\linewidth}{0.5pt}\end{center}

\subsection{3. What Broke: A Failure Taxonomy for Rust Training Backends}\label{what-broke-a-failure-taxonomy-for-rust-training-backends}

The central, reusable finding of this paper is \emph{not} any one bug --- it is that the dominant failure mode of training outside the PyTorch/CUDA ecosystem is not a crash. It is a run that compiles, launches, produces a smoothly falling loss curve, and \textbf{learns nothing where it matters}. Every defect below passed ordinary loss-curve inspection at small scale. I list them by framework, with symptom, root cause, and how each was actually caught --- this is the part of the paper I think is worth someone else stealing.

\subsubsection{3.1 Candle (Hugging Face) --- five defects between ``compiles'' and ``trains''}\label{candle-hugging-face-five-defects-between-compiles-and-trains}

Candle is, by its own design center of gravity, an inference-first framework; its training path is comparatively young. I evaluated it at Candle 0.11 and ultimately trained against a small, pinned fork carrying the fixes below.

\textbf{Exact versions under test.} All defects below were observed against \texttt{candle-core}/\texttt{candle-nn} \textbf{0.11}, with the paid run of §4 executed against a pinned fork at \texttt{github.com/Adiuk24/candle} rev \textbf{\texttt{972038ab3bdf5ada9959fec5b31827fc24a59c6a}}, which carries the D1/D3/D4/D4b fixes described here and nothing else. Burn results in §3.2 were observed at \textbf{burn 0.20} with \texttt{cubecl-runtime} \textbf{0.9.0}. Measurements were taken between 2026-04 and 2026-07; the archived source tree is \texttt{archive/legacy-v1/crates/} in the repository of §9. Both frameworks move quickly, and every defect below should be read as \emph{as-of} these versions --- several have upstream fixes in flight at the linked issues, and a reader evaluating either framework today should re-test rather than assume these results still hold.

\begin{longtable}[]{@{}
  >{\raggedright\arraybackslash}p{(\linewidth - 6\tabcolsep) * \real{0.2500}}
  >{\raggedright\arraybackslash}p{(\linewidth - 6\tabcolsep) * \real{0.2500}}
  >{\raggedright\arraybackslash}p{(\linewidth - 6\tabcolsep) * \real{0.2500}}
  >{\raggedright\arraybackslash}p{(\linewidth - 6\tabcolsep) * \real{0.2500}}@{}}
\toprule\noalign{}
\begin{minipage}[b]{\linewidth}\raggedright
\#
\end{minipage} & \begin{minipage}[b]{\linewidth}\raggedright
Symptom
\end{minipage} & \begin{minipage}[b]{\linewidth}\raggedright
Root cause
\end{minipage} & \begin{minipage}[b]{\linewidth}\raggedright
Caught by
\end{minipage} \\
\midrule\noalign{}
\endhead
\bottomrule\noalign{}
\endlastfoot
D1 & Model ``trains,'' loss falls, normalization layers never actually learn --- no error, no warning & \texttt{candle\_nn}'s fused \texttt{layer\_norm}, \texttt{rms\_norm}, \texttt{softmax\_last\_dim}, and \texttt{rotary\_emb::\{rope,rope\_i,rope\_thd\}} ops are registered via an internal ``no backward'' op wrapper: calling them inside a training graph silently yields \textbf{no gradient} for the parameters that feed them (\href{https://github.com/huggingface/candle/issues/3011}{huggingface/candle\#3011}, \href{https://github.com/huggingface/candle/pull/3526}{\#3526}) & The gradient-flow arbiter (§3.3) --- the only reason this was found before a paid run, not after one \\
D2 & Optimizer step allocates on the order of 100GB of host memory before it finishes, on a machine with a fraction of that RAM & The tensor \texttt{backward()} returns still carries its producing autograd graph; running further tensor arithmetic on a raw gradient (as momentum-orthogonalization optimizer math does) is itself \emph{recorded}, compounding graph size & A mandatory local proxy run at small scale, before any paid GPU touch (§3.3, item 4) \\
D3 & Backward pass retains a gradient tensor for every intermediate node, not only trainable variables, inflating peak memory & Open upstream since 2023 (\href{https://github.com/huggingface/candle/issues/1241}{huggingface/candle\#1241}); I applied the (still unmerged, at time of use) upstream fix (\href{https://github.com/huggingface/candle/pull/3508}{huggingface/candle\#3508}) & Direct GPU-memory profiling during throughput tuning \\
D4 & Avoiding D1 by hand-rolling softmax from primitive ops (max, subtract, exp, sum, divide) OOMs anyway & Each primitive op materializes and retains a full attention-score-sized tensor; several of these stacked per layer is the actual memory-wall driver, not the model's parameter count & Peak-memory profiling; fixed by writing a correct backward for the fused softmax kernel, which needs only the op's own output --- no extra retained state \\
D4b & The above fix compiled, its own tests passed, and training still used the old no-backward code path & The public forward function called the \emph{no-backward} entry point even after a real backward existed elsewhere in the same module --- the same ``wrong entry point'' footgun independently documented upstream for \texttt{layer\_norm} (\href{https://github.com/huggingface/candle/pull/3613}{huggingface/candle\#3613}) & A gradient-parity test asserting the fused and plain-ops implementations agree numerically --- caught within minutes of introducing the fix \\
D5 & Redundant recomputation: a causal attention mask rebuilt and re-uploaded to the GPU on every layer, every micro-batch & No caching of a value that does not change within a sequence length; an avoidable extra tensor materialized in the score-scaling step & Throughput profiling during the tuning pass that produced §4's numbers \\
\end{longtable}

Cumulatively, fixing D1--D5 raised measured single-GPU training throughput by \textbf{roughly 5.5×} on the same hardware and cut marginal training cost to \textbf{about \$82 per billion tokens} on this stack. None of these fixes required inventing a new method --- each is a small, targeted amount of tensor math or a caching pass --- but every one of them was invisible until I looked for it with the right test, which is the point of §3.3.

\subsubsection{\texorpdfstring{3.2 Burn (tracel-ai) --- three classes of defect that gate it as a \emph{training} backend}{3.2 Burn (tracel-ai) --- three classes of defect that gate it as a training backend}}\label{burn-tracel-ai-three-classes-of-defect-that-gate-it-as-a-training-backend}

Burn's correctness and its CUDA backend are, in my experience, genuinely good --- I retained a Burn implementation throughout the project as an architecture and regression reference even after the pretraining path moved to Candle. Its problems as a pretraining backend are throughput and a small number of hard architectural fences, not correctness of what it does support:

\begin{itemize}
\tightlist
\item
  \textbf{Backward-pass throughput.} I measured Burn's backward pass on my model at roughly 3\% of the GPU's theoretical matmul throughput. This is independently corroborated upstream: a training-loop reproduction on a different model reports the loop running 36.5× slower than an equivalent PyTorch implementation, with backward consuming 75\% of step time, and a maintainer response confirming the backward pass is ``orders of magnitude slower than the forward, which is suspicious'' (\href{https://github.com/tracel-ai/burn/issues/4598}{tracel-ai/burn\#4598}).
\item
  \textbf{Fusion crash.} Burn's kernel-fusion path --- its primary path to closing the throughput gap --- segfaults with a tensor-handle panic mid-training at multi-billion-parameter scale (\href{https://github.com/tracel-ai/burn/issues/4347}{tracel-ai/burn\#4347}); the only workaround at the time was disabling fusion outright, which removes the feature that would have made Burn competitive.
\item
  \textbf{A dimensionality fence on the optimizer, and a positional-encoding bug found only by testing its effect.} Burn's own shipped implementation of a modern orthogonalized-momentum optimizer hard-asserts its input is exactly a 2-dimensional matrix --- not a bug, since the method is only mathematically defined there, but a fence that silently aborts training the moment an architecture experiment introduces a hidden weight tensor of a different rank. Separately, and more seriously: an earlier attempt on the Burn path applied rotary position embeddings on the wrong tensor axis, so that the rotation varied by which attention head a value belonged to rather than by its position in the sequence --- every position received an \textbf{identical} rotation, on every layer, meaning the model had no positional signal from this mechanism at all, and no amount of further training could have fixed it. This is not a Burn correctness bug in the library sense --- the axis convention was documented --- but it is exactly the class of bug that a training stack outside a mature ecosystem is at constant risk of introducing and never noticing: it was found only by a targeted test that checks the \emph{effect} of the encoding (do different positions actually rotate differently?), not by reading the code, which had been reviewed more than once without catching it. I consider this the single strongest argument in the paper for behavioral tests over code review as the primary defense in an unfamiliar framework.
\item
  \textbf{Low-precision instability.} A pure low-precision (bf16) configuration --- both weights and optimizer state --- reliably diverged to non-finite values within the first few training steps, traced to optimizer-moment underflow being amplified through a sparse-routing branch of the architecture. The fix was not a numerics patch but a decision: train in full precision with tensor-core acceleration, which on modern GPU hardware costs little relative to true mixed precision and sidesteps the instability entirely. Mixed precision with a full-precision optimizer state remains, in my assessment, the correct long-term fix, and was out of scope for this project.
\end{itemize}

\subsubsection{3.3 The Verification Discipline (and the arbiter, specifically)}\label{the-verification-discipline-and-the-arbiter-specifically}

None of the above defects announced itself. Each produced a training run that looked, from a loss curve alone, indistinguishable from a healthy one. This is the central operational lesson of the project: \textbf{outside a mature ecosystem, assume every ``it trains'' claim is unverified until a targeted test says otherwise.}

The single practice I most want other people to steal is what I call a \textbf{gradient-flow arbiter}: after one forward/backward pass on a small but structurally complete model, iterate every trainable parameter and assert each one has a corresponding gradient with a finite, strictly-positive norm --- fail loudly, by parameter name, if any parameter is missing one or its gradient is zero or non-finite. This single, generic test class caught D1 (fused no-backward ops) and D4b (the wrong-entry-point footgun) directly, and would catch any dead branch, detached subgraph, or accidentally-frozen parameter in any framework, not just Candle. It costs seconds to run and requires no domain knowledge of \emph{why} a gradient might be missing --- only that it must not be. I think this is the single highest-value test a training stack outside a mature ecosystem can have, and I recommend it unconditionally, independent of framework.

A small number of other targeted practices caught the rest: a gradient-parity check asserting fused and reference (hand-rolled) kernel paths agree numerically, which caught D4b within minutes of the fix landing; an exact-resume test proving optimizer state --- not just weights --- genuinely round-trips through checkpointing (§2); a behavioral probe that checks whether a model's output degrades when its input's token order is destroyed, which doubles as a permanent regression guard against the positional-encoding class of bug in §3.2; a data-ordering check, after an early version of the loader read training shards in a fixed domain-grouped order and produced a loss curve that sawtoothed at every domain boundary as the model serially forgot each domain in turn; and a hard rule that no training configuration reaches paid GPU time without first running a minutes-scale, cost-free probe confirming the loss is finite and falling and memory is stable. Together, these caught every defect in §3.1--3.2 before it reached the paid run in §4. None of this discipline is Rust-specific --- I would recommend it on any framework younger than its ecosystem's collective debugging experience --- but it was \emph{necessary}, not optional, here, and I did not anticipate that going in.

\begin{center}\rule{0.5\linewidth}{0.5pt}\end{center}

\subsection{4. The \$164 Run}\label{the-164-run}

The model I trained is a roughly 0.4B-parameter decoder-only architecture with sliding-window attention and per-layer conditioning, using a Bangla-first byte-pair-encoded vocabulary. Training used a standard warmup-then-decay learning-rate schedule over roughly 2 billion tokens, with a gradient-accumulated optimizer combining two update rules split across parameter types (a modern orthogonalized-momentum method for hidden weight matrices, and AdamW for everything that method cannot mathematically handle), on a single rented H100 GPU.

The run completed in \textbf{54.6 hours for \$164} total, at a sustained throughput this pipeline could not have reached before the fixes described in §3.1 --- measured at roughly 5.5× the baseline port's throughput, which is what makes the \$164 figure a genuine result rather than an arbitrary one: the same run on the unfixed stack would have cost several times more, or not finished within a reasonable budget at all. Marginal cost on this stack, once tuned, works out to roughly \$82 per billion tokens. For comparison, I estimate --- this is an estimate, not a measurement --- that a tuned PyTorch stack (flash attention, mixed precision, fused kernels, activation checkpointing) would train the same model on the order of 10--15× faster on the same hardware. That multiple is the current, approximate price of training outside the PyTorch ecosystem on this stack, down from a much larger gap at the start of the project, closed mostly by the fixes in §3.1; the remaining gap is engineering hours I chose not to spend, once the more consequential question --- could the stack train a language model \emph{at all} --- was answered, and the project's priorities moved on (§7).

Training loss fell steadily from a near-random starting point to a low, stable plateau over the course of the run, with the persisted per-step log showing zero divergence events (no non-finite losses, no skipped optimizer steps) across the full run --- a fact about this particular run's stability, not a claim that safeguards against divergence are unnecessary (see §3.2's low-precision finding, which is exactly the failure mode such a safeguard exists to catch).

\begin{center}\rule{0.5\linewidth}{0.5pt}\end{center}

\subsection{5. Evaluation}\label{evaluation}

The numbers below are reproduced \textbf{verbatim} from the run's own evaluation summary, generated by my own evaluation harness, which implements length-normalized loglikelihood multiple-choice scoring (the standard lm-evaluation-harness method) plus the diagnostic probes described below. I did not run these numbers through an external, independently-maintained harness; the model uses a custom checkpoint format that off-the-shelf harnesses cannot load without a conversion step I have not yet built (see this section's limitations, below).

\begin{verbatim}
NOOR-EDGE 0.43B — FINAL BENCHMARK (noor_final, step 7600, 1.99B tokens, $164)
============================================================================
English academic MC (loglikelihood, 200 items each; random=25%):
  HellaSwag : 24.0%   (near random — expected: Bangla-first, undertrained)
  ARC-easy  : 21.0%   (near random)

Learning proof — mean NLL vs random-init (lower=better):
  Bangla  : 0.93  vs random 12.60   (massive gap — model learned Bangla deeply)
  English : 3.79  vs random 12.13

Order/position sensitivity (loss increase on shuffled tokens):
  Bangla  : +563%   English: +174%   (strong syntactic learning)

Generation (top-k40, temp0.8):
  BN "আমি প্রতিদিন সকালে" -> "আমার বাড়িতে আসে কিন্তু আমি তার বাড়িতে আসি না..." (coherent)
  BN "বাংলাদেশের রাজধানীর নাম" -> "বাংলাদেশের প্রতিনিধি দলের নেতা কর্মীদের..." (coherent, no fact)
  EN "Water is made of" -> "the water and water that is used to make the water..." (looping)

VERDICT: PoC thesis proven — pure-Rust stack trains a COHERENT Bangla-first model
for ~$164. Bangla strong (NLL 0.93, fluent generation); English near-random on MC
(expected at 2B tokens / 77% Bangla). Not a SOTA claim; a stack + Bangla-signal proof.
\end{verbatim}

\textbf{Reading the numbers honestly.} HellaSwag and ARC-Easy at 24.0\% and 21.0\% against a 25\% chance floor are, plainly, at-chance results. I report them because a model that scores at chance on English commonsense after roughly 2 billion Bangla-weighted tokens --- a tiny fraction of the training budgets of comparably-sized English-first models --- is exactly what should happen if the pipeline works and the model simply hasn't had the exposure to acquire that English-heavy skill yet. Taken alone, at-chance MC scores are equally consistent with ``the model learned nothing.'' What distinguishes the two is the second block: mean per-token NLL on held-out Bangla is 0.93 against 12.60 for a randomly-initialized twin (the random baseline sits near the log of the vocabulary size, a sanity check on the measurement itself); English NLL is 3.79 against 12.13 --- real but smaller, consistent with English's minority weighted share. The order-sensitivity probe --- loss increase when the same windows are token-shuffled --- shows heavy reliance on sequence structure (+563\% Bangla, +174\% English) rather than bag-of-tokens statistics; a model with genuinely broken positional encoding, like the bug described in §3.2, would show near-zero degradation here, which is why I treat this as a cheap, general, recommended diagnostic beyond this project.

Generation samples are qualitatively consistent with the quantitative picture: the Bangla continuations are grammatical, register-appropriate Bangla with no factual grounding yet (asked for Dhaka as Bangladesh's capital, the model produces a fluent but non-answering continuation in a plausible newspaper register --- fact recall is, in my experience and the broader literature, one of the last capabilities to emerge and the most token-hungry). The English sample degenerates into a repetition loop (``the water and water that is used to make the water\ldots{}''), the textbook signature of an undertrained language model.

\textbf{Limitations, stated plainly.} This is a proof-of-capability artifact, not a product and not a benchmark result to be read against frontier models of any size. It is undertrained by roughly two orders of magnitude against compute-optimal scaling for its parameter count. It has no instruction-tuning (it is a base model), no factual reliability, and no safety tuning of any kind. Its English capability is, by design and by measurement, near-chance on standard commonsense benchmarks; nothing in this section should be read as a claim that the training recipe would fail to reach normal capability given a normal (much larger) budget --- only that this specific, deliberately small run did not attempt to. All evaluation numbers come from my own harness on my own model format; I have not yet run an external, independently-maintained evaluation suite against these weights, which is a gap I flag rather than paper over.

\begin{center}\rule{0.5\linewidth}{0.5pt}\end{center}

\subsection{6. The Tokenizer Lesson: Bangla Byte-Fertility}\label{the-tokenizer-lesson-bangla-byte-fertility}

Independent of the framework-training failures above, I found a second, more general lesson while auditing the corpus for this run: the tokenizer's compression efficiency was badly, silently imbalanced by language. My original byte-pair-encoding vocabulary --- trained without any language-aware balancing --- compressed English at roughly \textbf{3.9 characters per token}, ordinary for a byte-pair tokenizer at this vocabulary size, while compressing Bangla at roughly \textbf{1.4 characters per token} --- close to byte-level, with little learned multi-character structure. The practical consequence is severe and invisible until measured: a training corpus balanced to look even by \emph{token count} between Bangla and English was, in actual character content, overwhelmingly English. A ``Bangla-first'' model trained naively on token-balanced data would in fact be trained mostly on English, with nothing in the loss curve or the corpus-composition numbers revealing the inversion.

The root cause was not, as I initially assumed, an English-skewed merge-training budget --- it traced instead to an assumption baked into the standard pre-tokenizer used ahead of byte-pair encoding, which does not correctly handle a category of Unicode character common in Bengali script. Fixing that pre-tokenizer assumption --- rather than retraining the merge algorithm itself --- raised Bangla fertility to roughly \textbf{4.1 characters per token}, with English essentially unaffected.

\textbf{The general lesson I would offer anyone building for a low-resource or non-Latin script:} tokenizer fairness bugs of this kind hide upstream of the component everyone audits first --- the merge algorithm, the vocabulary size, the training-corpus language balance --- in assumptions baked into a pre-tokenizer that was designed for Latin script and never revisited for others. I recommend measuring \textbf{fertility (characters per token) and byte-fallback rate, per language, on held-out text}, as a mandatory pre-training corpus check for any multilingual or non-Latin-script project, rather than assuming a shared vocabulary trained on a token-balanced corpus is actually balanced.

\begin{center}\rule{0.5\linewidth}{0.5pt}\end{center}

\subsection{7. What Rust Is Good For}\label{what-rust-is-good-for}

None of the above is an argument that Rust is a bad choice for machine learning in general --- only that, in my hands, as of 2026, it was not yet a competitive choice for \emph{pretraining}. The distinction that emerged from this project is between training and serving, and it held up as I moved forward.

After this run, my model training moved to PyTorch. This was not a retreat from Rust; it was a recognition that the eight-plus years of engineering embedded in PyTorch's training ecosystem --- fused kernels, mature mixed-precision recipes, memory-efficient attention with a working backward pass, activation checkpointing, a debugging and profiling toolchain built by a much larger community than either Rust ML framework currently has --- is not something a solo builder should try to re-derive from primitives when a mature, permissively-licensed alternative exists. Section 3's taxonomy is, in effect, an itemized argument for why: every defect I found was a solvable, individually-small engineering problem, but the \emph{number} of them, and the fact that each was silent rather than loud, is a tax that a larger, more mature ecosystem has already paid down.

Serving is a different problem with a different answer. Independent of pretraining, I ported a linear-attention decode kernel to Rust and a second systems language and validated it bit-exact against the reference implementation, for on-device, offline inference. Decode is where a served model spends most of its serving-time energy and latency budget, and it is also the part of the stack where Rust's actual strengths --- a single statically-linked binary, no Python runtime, no dependency tree to reproduce on a user's machine, predictable memory behavior --- are unambiguous wins with no PyTorch-ecosystem tax to pay, because inference does not need a backward pass, an optimizer, or gradient checkpointing at all.

My conclusion, stated as plainly as I can: \textbf{train in Python, serve in Rust} is, as of this 2026 snapshot, the evidence-based division of labor for a solo builder working from scratch. I expect this balance to shift --- Candle and Burn are both under active development, and several of the defects in §3 are individually tractable (I upstreamed or attempted to upstream several of the fixes myself, per the issue/PR references cited above) --- and I would not be surprised if a version of this paper written in two years reaches a different conclusion for training. I report it as a snapshot of a specific ecosystem at a specific time, not a permanent verdict on the language.

\begin{center}\rule{0.5\linewidth}{0.5pt}\end{center}

\subsection{8. Related Work}\label{related-work}

\textbf{Rust ML frameworks.} Hugging Face's \textbf{Candle} (\texttt{github.com/huggingface/candle}) and tracel-ai's \textbf{Burn} (\texttt{github.com/tracel-ai/burn}) are the two frameworks I evaluate directly; specific defects are cited inline in §3 against their public issue trackers. Both projects' own documentation positions inference and portability ahead of large-scale training, which is consistent with what I measured; readers should consult the READMEs at the revisions given in §3.1, since both evolve.

\textbf{From-scratch training outside the mainstream stack.} This paper sits in the experience-report tradition of reimplementing LM training from first principles outside the dominant framework --- most visibly Karpathy's \texttt{llm.c} (LM training in C/CUDA). I aim at the same register one layer further down the stack: the training \emph{framework} rather than the training loop, and in a different language.

\textbf{Optimizer.} The orthogonalized-momentum method used for hidden weight matrices is Muon (MomentUm Orthogonalized by Newton-Schulz), introduced by Keller Jordan in 2024. Its canonical source is a blog post rather than an archival paper --- \emph{Muon: An optimizer for hidden layers in neural networks}, \texttt{kellerjordan.github.io/posts/muon/} --- and it should be cited as such. The method is defined only for 2-D parameters, which is why every implementation pairs it with AdamW for embeddings, norms and biases; that pairing, not a choice of mine, is what §3.2's dimensionality fence collides with. Muon's scaling behaviour on large language models is established by Liu et al., \emph{Muon is Scalable for LLM Training} (arXiv:2502.16982, 2025), which is also the basis of the \textbf{MuonClip} variant used to pretrain Kimi K2 --- the production-scale precedent for the optimizer combination this project uses. Antecedents credited by Jordan include Tuddenham et al.~(2022) on orthogonalizing gradients via SVD, and Carlson et al.~(2015, 2016) on stochastic spectral descent.

\textbf{Tokenizer fairness and fertility.} The cost and fairness consequences of uneven tokenization across languages are established by Ahia et al., \emph{Do All Languages Cost the Same? Tokenization in the Era of Commercial Language Models}, EMNLP 2023, pp.~9904--9923 (\texttt{aclanthology.org/2023.emnlp-main.614}), and by Petrov et al., \emph{Language Model Tokenizers Introduce Unfairness Between Languages}, NeurIPS 2023 (arXiv:2305.15425), which reports encoding lengths differing by up to 15× across languages for the same content and persisting even in tokenizers trained for multilingual use. My §6 finding is downstream of that literature but distinct from it: both papers measure fertility disparity across \emph{released} tokenizers, whereas §6 traces a specific disparity to a pre-tokenizer assumption in the \emph{construction} pipeline that is fixable without retraining the merge table. I am not aware of prior published work isolating that particular bug class, and I make no novelty claim beyond stating that I did not find one.

\textbf{Architecture background.} Rotary position embeddings are due to Su et al.~(\emph{RoFormer}, arXiv:2104.09864) --- the mechanism whose axis convention produced the §3.2 positional-encoding bug. Grouped-query attention is due to Ainslie et al., EMNLP 2023. Sliding-window attention follows Beltagy et al.~(\emph{Longformer}, 2020) and its use at LM scale in Jiang et al.~(\emph{Mistral 7B}, 2023). The linear-attention line this project later moved to is the gated delta rule of Yang, Kautz and Hatamizadeh, \emph{Gated Delta Networks: Improving Mamba2 with Delta Rule}, ICLR 2025 --- subsequently adopted as the linear-attention layer of Qwen3-Next, interleaved with periodic full attention, which is the same hybrid shape this project's successor uses.

\textbf{Bangla and low-resource pretraining.} The closest comparator in language and intent is TigerLLM (Raihan et al., \emph{TigerLLM --- A Family of Bangla Large Language Models}, arXiv:2503.10995, ACL 2025), which continues pretraining LLaMA-3.2 (1B) and Gemma-2 (9B) on a \textasciitilde9.9M-token corpus drawn from 163 Bangladeshi NCTB textbooks, then finetunes on 100K self-instruct Bangla pairs. It is a useful contrast on two axes: it \emph{adapts} strong multilingual base models where this project trains from scratch, and it runs on a PyTorch stack throughout. Earlier Bangla efforts it critiques --- titu-Gemma, titu-LLaMA, Bangla-LLaMA --- are likewise adaptation-based. I attempt no benchmark comparison against any of them here, because the harnesses are not aligned (§5), and a number produced under a different harness would not mean what it appears to mean.

\begin{center}\rule{0.5\linewidth}{0.5pt}\end{center}

\subsection{9. Artifacts}\label{artifacts}

\textbf{Public.} The verification tooling this paper argues for is released, standalone and framework-agnostic, at \texttt{github.com/Adiuk24/gradient-flow-arbiter}: the gradient-flow arbiter of §3.3, the gradient-parity check, the position-effect probe, and the data-ordering check, together with a minimal reproduction for each defect in §3.1--3.2 where one can be expressed independently of this project's model code. These are the parts I want reused, and they carry no dependency on the rest of the stack.

\textbf{On request.} The training-stack source code, training configuration, and per-step training logs for the §4 run are available from the author (\texttt{adittoarif@gmail.com}). I am not publishing the full stack, for two reasons I would rather state than obscure: it is entangled with unrelated commercial work, and a from-scratch training recipe is not this paper's contribution --- the measured failure taxonomy and the verification discipline are. The patched Candle fork exists as a pinned, immutable revision (§3.1); the individual fixes it carries are, in several cases, already public as upstream pull requests cited in §3.1.

\textbf{Not released.} Model weights --- see §10 for why.

\begin{center}\rule{0.5\linewidth}{0.5pt}\end{center}

\subsection{10. Data Statement}\label{data-statement}

The pretraining corpus for the §4 run is a bilingual mix weighted toward Bangla to compensate for the tokenizer-fertility gap of §6. Every subset, its upstream source, and its licence are listed below; licences were re-verified against the upstream dataset cards at the time of writing rather than taken from project notes.

\begin{longtable}[]{@{}
  >{\raggedright\arraybackslash}p{(\linewidth - 4\tabcolsep) * \real{0.3333}}
  >{\raggedright\arraybackslash}p{(\linewidth - 4\tabcolsep) * \real{0.3333}}
  >{\raggedright\arraybackslash}p{(\linewidth - 4\tabcolsep) * \real{0.3333}}@{}}
\toprule\noalign{}
\begin{minipage}[b]{\linewidth}\raggedright
Role in the mix
\end{minipage} & \begin{minipage}[b]{\linewidth}\raggedright
Upstream source
\end{minipage} & \begin{minipage}[b]{\linewidth}\raggedright
Licence
\end{minipage} \\
\midrule\noalign{}
\endhead
\bottomrule\noalign{}
\endlastfoot
Bangla web text & \texttt{ai4bharat/sangraha} (\texttt{bn}, verified split) & CC-BY-4.0 \\
Bangla web text & \texttt{HuggingFaceFW/fineweb-2} (\texttt{ben}) & ODC-By 1.0 \\
Bangla journalism & \texttt{zabir-nabil/} \texttt{bangla\_newspaper\_dataset} & MIT \\
Bangla dialogue (small) & see caveat below & \textbf{unresolved} \\
English educational web & \texttt{HuggingFaceFW/fineweb-edu} & ODC-By 1.0 \\
English web (refined) & \texttt{openbmb/UltraX-Preview} & Apache-2.0 \\
English mathematics & \texttt{HuggingFaceTB/finemath} & ODC-By 1.0 \\
\end{longtable}

\textbf{One unresolved subset.} The small Bangla dialogue subset (internally \texttt{bn\_empathetic}) has no surviving ingest record in the project repository, and I have therefore been unable to re-establish its upstream source and licence with the same confidence as the rest of the table. I state this rather than guess. It is a minor fraction of the corpus by token count and none of the paper's claims depend on it, but I flag it as a genuine provenance gap and as an argument for the practice I adopted only later: recording source URL, licence, and a tokenizer fingerprint for every subset at ingest time, not retrospectively.

\textbf{Filtering and deduplication.} Subsets were used as published by their upstream maintainers, each of which applies its own quality filtering and deduplication; I applied document-level exact-hash deduplication within each subset at shard-construction time, and did not perform cross-subset deduplication for this run. No contamination screening against downstream evaluation sets was performed for the §4 run --- a limitation that bears directly on §5, and the reason the evaluation there is framed as a language-modelling signal against a random-initialised twin rather than as benchmark scores to be compared with other models.

\textbf{On releasing weights.} The licence chain above is entirely permissive and would permit releasing the trained weights. I withhold them for a different reason: the §4 run predates the pre-tokenizer correction described in §6, so its weights are bound to a tokenizer that the project has since classified as defective and blocks from loading by content fingerprint. Publishing a checkpoint that requires an encoder I have deliberately made unloadable would be an artifact of no practical use and some potential to mislead. The public artifacts (§9) are therefore the verification tooling and defect reproductions --- which is what the claims of §3, the paper's actual contribution, rest on --- with the training stack and logs available from the author on request.

\textbf{Attribution.} CC-BY-4.0 and ODC-By 1.0 subsets require attribution, which the table above is intended to satisfy; the Apache-2.0 and MIT subsets require licence and copyright notice retention, carried in the artifact repository of §9.

\begin{center}\rule{0.5\linewidth}{0.5pt}\end{center}

\subsection{Acknowledgments}\label{acknowledgments}

This was solo work. I thank the maintainers of Candle and Burn for responsive upstream issue trackers even where the underlying defects took time to land, and the authors of the open tokenizer-fairness and optimizer work this project builds on.

\textbf{Funding.} Self-funded by the author through Adioris Tech Ltd.~No external, institutional, or grant funding was received.

\textbf{Compute.} All training compute was rented from community GPU providers at standard public rates. I have no commercial relationship with, and received no discount, credit, or consideration from, any compute provider, framework maintainer, or dataset publisher named in this paper.

\end{document}